\documentclass[11pt]{article}

\usepackage[preprint]{acl}

\usepackage{times}
\usepackage{latexsym}

\usepackage[T1]{fontenc}

\usepackage[utf8]{inputenc}

\usepackage{microtype}

\usepackage{inconsolata}

\usepackage{graphicx}

\usepackage{multirow}
\usepackage{bbding} 
\usepackage{amsmath}
\usepackage{amssymb}
\usepackage{algorithm}
\usepackage{algorithmic}
\usepackage{longtable} 

\title{GeoChallenge: A Multi-Answer Multiple-Choice Benchmark for Geometric Reasoning with Diagrams}

\author{
	\textbf{Yushun Zhang\textsuperscript{1,2}},
	\textbf{Weiping Fu\textsuperscript{1,2}},
	\textbf{Zesheng Yang\textsuperscript{1,2}},
	\textbf{Bo Zhao\textsuperscript{1,2}},
	\textbf{Lingling Zhang\textsuperscript{1,2}},
	\\
	\textbf{Jian Zhang\textsuperscript{1,2}},
	\textbf{Yumeng Fu\textsuperscript{1,2}},
	\textbf{Jiaxing Huang\textsuperscript{1,2}},
	\textbf{Jun Liu\textsuperscript{1,2}},
	\\
	\\
	\textsuperscript{1}School of Computer Science and Technology, Xi’an Jiaotong University, Xi’an, China,
	\\
	\textsuperscript{2}Shaanxi Province Key Laboratory of Big Data Knowledge Engineering, Xi’an, China
	\\
}

\begin{document}
	\maketitle
	\begin{abstract}
		
		Evaluating the symbolic reasoning of large language models (LLMs) calls for geometry benchmarks that require multi-step proofs grounded in both text and diagrams. However, existing benchmarks are often limited in scale and rarely provide visually grounded multiple-choice questions, limiting reliable evaluation of complex reasoning. We introduce GeoChallenge, a dataset of 90K automatically generated multiple-choice geometry proof problems, each requiring multi-step reasoning over aligned textual descriptions and diagrams. GeoChallenge provides fine-grained complexity ratings and formal language annotations to enable controlled evaluation. 
		
		Experiments on multiple advanced LLMs show a clear performance gap between models and humans (the best-performing model, GPT-5-nano, achieves 75.89 exact match vs. 94.74 for humans). Further analysis also reveals three common failure patterns of LLMs: (1) exact match failures under the multiple-choice setting; (2) weak visual reliance; and (3) overextended reasoning without convergence.\footnote{Codes and resources will be available at: \url{https://github.com/fanhualiushang/GeoChallenge}}

	\end{abstract}
	
	\section{Introduction}
	Geometry problem solving is a fundamental task involving spatial reasoning and symbolic deduction \cite{trinh2024solving,alphageometry2,pan2025enhancinggeometricproblemsolvingability,dai2025FromSymbolicPerceptiontoLogicalDeduction}, and is widely used to evaluate complex reasoning in large language models (LLMs) \cite{luo2025geogrambenchbenchmarkinggeometricprogram,feng2025geobenchrethinking,zhang2025gkgllmunifiedframeworkgeneralized}. Yet evaluating such capabilities of LLMs in a systematic and scalable manner remains an open challenge \cite{s3eval,hariharan2025breakpointscalableevaluationsystemlevel,QuantifyingDataContamination,white2025livebenchchallengingcontaminationlimitedllm, Geomverse}.
	
	\begin{figure}[t]
		\centering
		\includegraphics[width=\columnwidth]{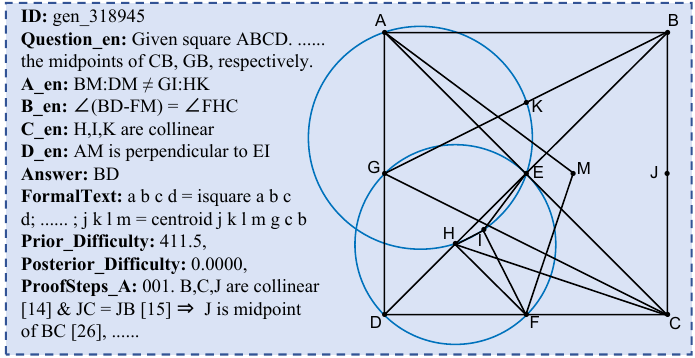}
		\caption{Examples in GeoChallenge-90K dataset.}
		\label{figure:example_problem}
	\end{figure}
	
	Geometry datasets and benchmarks have progressed along two complementary lines. Early efforts such as Geometry3K~\cite{InterGPS} and GeoQA~\cite{GeoQA} translate diagram--text problems into machine-interpretable structures (e.g., symbolic relations or executable programs), enabling systematic evaluation of geometric reasoning. More recent benchmarks, including MathVerse~\cite{zhang2024mathverse} and OlympiadBench~\cite{he2024olympiadbench}, target multimodal ~\cite{zhang2025mapsmultiagentpersonalityshaping} and emphasize faithful diagram understanding in harder visual-math settings. However, reliable evaluation remains challenging: many datasets are manually curated from textbooks or contests, which limits coverage and scalability, and high-difficulty benchmarks stay small due to expert verification costs. Moreover, prevailing single-answer multiple-choice~\cite{feng2025geobenchrethinking,wang2025geomrel,xing2024gepbench} or open-ended formats~\cite{zhang2025marsmultiagentadaptivereasoning,fu2025geolaux} are either prone to guessing or difficult to grade at scale. Finally, text--diagram misalignment persists, weakening visually grounded evaluation, as highlighted by GeoGPT4V~\cite{Geogpt4v}.

	To address these issues, we propose a scalable automatic generation pipeline and introduce GeoChallenge-90K, a dataset of 90,279 challenging geometry proof questions. Each instance provides an aligned textual description, a rendered diagram, and four candidate options with possibly multiple correct answers, enabling option-level evaluation and discouraging elimination-style guessing. GeoChallenge-90K covers a wide range of proof length (avg. 16.72 steps), diagram complexity, and difficulty, with fine-grained complexity ratings to support controlled stress tests. 
	
	
	Experiments on GeoChallenge-90K reveal a large gap between current models and human solvers. General-purpose models average 21.48\% accuracy, reasoning-oriented models reach 56.07\%, and humans achieve 94.74\%. Hierarchical evaluation shows that general-purpose models degrade steeply with increasing complexity, whereas humans and reasoning-specialized models remain comparatively stable. Diagram ablations expose a grounding gap: removing diagrams substantially reduces human accuracy but only marginally affects models, suggesting that current LLMs do not reliably extract or calibrate diagram evidence. Error analysis further indicates recurring failures-logical fallacies, invalid outputs, and overextended reasoning with no verifiable conclusion-while human mistakes are mostly genuine reasoning slips rather than format-related failures.
	
	In summary, our main contributions are:
	\begin{itemize}
		\item We introduce GeoChallenge-90K, a multi-answer multiple-choice benchmark for diagram-grounded geometric reasoning, with aligned text--diagram pairs, formal annotations, and fine-grained complexity control.
		\item Extensive experiments provide systematic empirical evidence that a substantial model–human gap persists on challenging, long-step geometric reasoning, even under rigorous, option-level no-guess evaluation.
		\item Diagnostics reveal large gaps between strict exact match and option-level metrics, weak and inconsistent diagram grounding, and frequent answer inconsistency or non-convergent long-step reasoning.
		
	\end{itemize}
	
	\section{Related Work}
	\subsection{Geometry Problem Generation}
	
	Early geometry datasets were built via templates or manual collection andannotation~\cite{prasetyanto2020automatic, Geomverse, Mavis, GeoQA, InterGPS, Unigeo}, which produced valuable resources but faced scalability and coverage limits. Symbolic generation alleviates these issues by making problems machine-verifiable: Inter-GPS~\cite{InterGPS} provides parsing/representation, GeoGen~\cite{GeoGen} generates symmetric instances, FormalGeo~\cite{FormalGeo} formalizes verification, and R-CoT~\cite{R-cot} increases QA diversity. More recently, theorem-guided and verified pipelines further scale generation and alignment, including TR-CoT~\cite{deng2024theorem}, GenesisGeo~\cite{zhu2025genesisgeo}, and TrustGeoGen~\cite{fu2025trustgeogen}.
	
	LLMs/MMs introduce another paradigm. G-LLaVA~\cite{Gllava} synthesizes Geo170K with text LLMs, while GeoGPT4V~\cite{Geogpt4v} leverages GPT-4V/Wolfram to improve difficulty and image--text alignment, demonstrating scalable, targeted generation~\cite{chen2024premise, li2024synthesize}.
	
	\subsection{Existing geometry datasets}
	
	Geometry benchmarks range from early text--diagram datasets (e.g., GeoS~\cite{geos}) to large-scale, richly annotated corpora such as Geometry3K~\cite{InterGPS}, GeoQA/GeoQA+~\cite{GeoQA,GeoQA+}, UniGeo~\cite{Unigeo}, and PGPS9K~\cite{zhang2023pgps9k}, which provide formal languages, programmatic solutions, and detailed diagram annotations. GeoLaux~\cite{fu2025geolaux} further targets long-step reasoning with auxiliary-line constructions.
	
	With LLMs/MMs, benchmarks increasingly stress multimodal understanding, robustness, and perception, including MATH~\cite{hendrycks2021MATH}, GeoEval~\cite{zhang2024geoeval}, MM-MATH~\cite{sun2024Mmmath}, GePBench~\cite{xing2024gepbench}, and FrontierMath~\cite{glazer2024frontiermath}.
	
	\section{The GeoChallenge-90K dataset}
	
	
	\begin{figure*}[!t]
		\centering
		\includegraphics[width=0.99\textwidth]{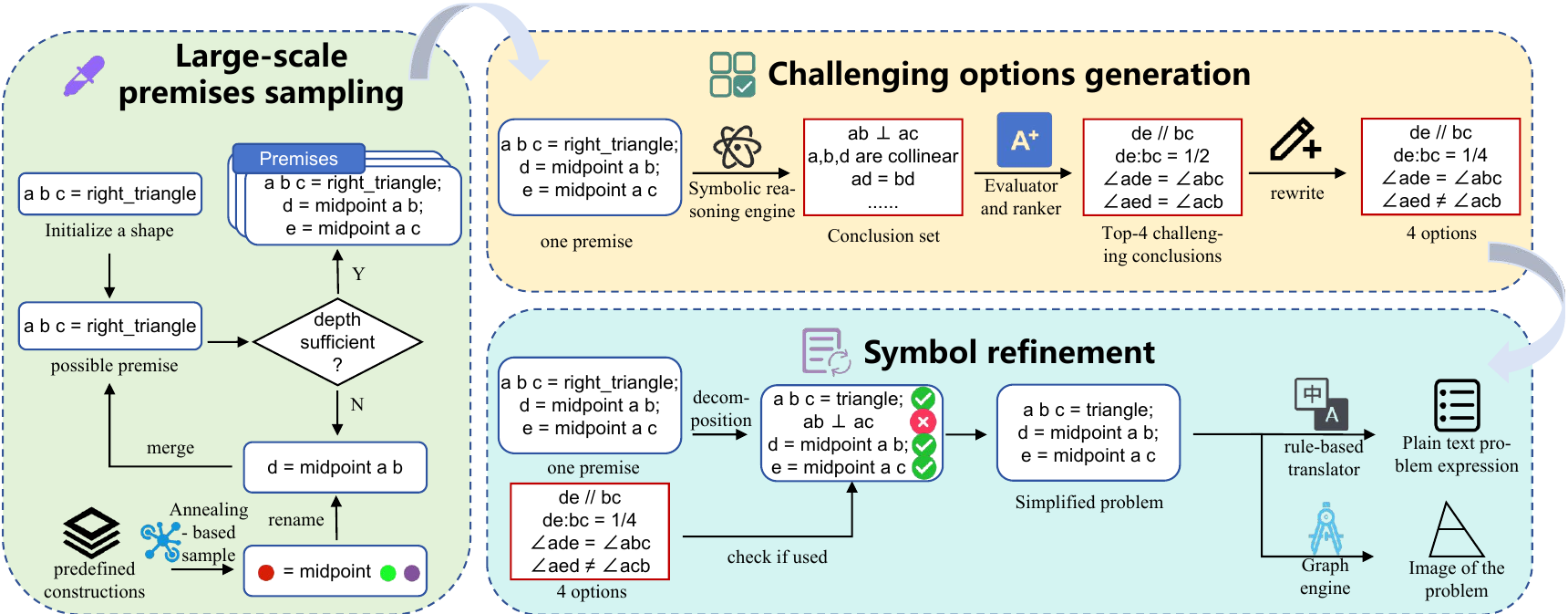}
		\caption{Pipeline of dataset generation}
		\label{figure:pipeline}
	\end{figure*}
	
	\subsection{Definitions}
	
	We first introduce the core concepts used in our automatic data generation procedure. Concrete examples are provided in Appendix~\ref{sec:reference_examples}.
	
	\paragraph{Clause.}
	A clause is a statement that specifies a geometric relation among a set of points and existing objects (e.g., points, lines, circles):
	\[
	f(X_1, X_2, \dots, X_n),
	\]
	where $f$ is a predefined relation and $X_1$, $X_2$, $\dots$, $X_n$ are already-defined points. All predefined clauses are listed in table ~\ref{table:defs_template}.
	
	\paragraph{Construction.}
	A construction is a geometric description that uses one or two clauses to uniquely define a new point $x$; when two clauses are given, $x$ is taken as their intersection:
	\[
	Construction\ x = f(), g(),
	\]
	where $f$ and $g$ are clauses, and $g()$ may be omitted if $f()$ alone uniquely determines $x$.
	
	\paragraph{Premise.}
	A premise is an ordered sequence of constructions that defines all points and relations in a problem:
	\[
	\begin{aligned}
		Premise = (&Construction_1, Construction_2, \\
		& \ldots, Construction_n).
	\end{aligned}
	\]
	\paragraph{Problem.}
	A problem is a structured task or question, which consists of a premise and four multiple-choice options (possibly with multiple correct answers):
	\begin{equation}
		\begin{aligned}
			Problem = (&Premise, Option_1, Option_2, \\
			&Option_3, Option_4).
		\end{aligned}
	\end{equation}
	See Figure~\ref{figure:example_problem} for an example.
	
	\subsection{Problem Generation}
	\label{section:problem_generation}
	

	\subsubsection{Large-scale premises sampling}
	
	We generate a large pool of premises by composing predefined clause templates into multi-layer geometric constructions. Starting from a minimal seed at depth $0$, we iteratively add layers until a target depth is reached. At each layer, we sample one or two templates and instantiate them with valid arguments from currently defined entities, ensuring consistency, then append the resulting clauses to the premise.
	
	To balance diversity and tractability, we perform breadth-first expansion with a gradually shrinking branching factor (annealing-style): early layers explore more template/parameter combinations, while later layers prune candidates to avoid combinatorial explosion. We set the maximum depth to $N=8$, producing 871,828 sampled premises.
	

	\begin{table*}[!t]
		\centering
		\small
		\begin{tabular}{cccccccccc}
			\hline
			\textbf{Dataset} & \textbf{Size} & \textbf{AG} & \textbf{IF} & \textbf{CR} & \textbf{Avg PL} & \textbf{Avg DL} & \textbf{FA} & \textbf{LT} & \textbf{QT} \\ \hline
			GeoS~\cite{geos} & 186 & \XSolidBrush & $T, I$ & \XSolidBrush & NA & 23.34 & \XSolidBrush & EN & SA \\
			MATH~\cite{hendrycks2021MATH} & 12,500 & \XSolidBrush & $T$ only & \CheckmarkBold & 5.48 & 202.66 & \XSolidBrush & EN & OE \\
			Geometry3K~\cite{InterGPS} & 3,002 & \XSolidBrush & $T, I$ & \XSolidBrush & NA & 12.19 & \CheckmarkBold & EN & SA \\
			GeoQA~\cite{GeoQA} & 5,010 & \XSolidBrush & $T, I$ & \XSolidBrush & 2.77 & 52.46 & \XSolidBrush & ZH & SA \\
			UniGeo~\cite{Unigeo} & 9,543 & \XSolidBrush & $T, I$ & \XSolidBrush & 2.57 & 16.23 & \CheckmarkBold & EN & SA, OE \\
			GPSM8K~\cite{cobbe2021GSM8K} & 8,790 & \XSolidBrush & $T$ only & \XSolidBrush & 4.59 & 45.27 & \XSolidBrush & EN & OE \\
			PGDP5K~\cite{hao2022pgdp5k} & 5,000 & \XSolidBrush & $I$ only & \XSolidBrush & -- & -- & \CheckmarkBold & -- & -- \\
			PGPS9K~\cite{zhang2023pgps9k} & 9,022 & \XSolidBrush & $T, I$ & \XSolidBrush & 2.43 & 9.46 & \CheckmarkBold & EN & OE \\
			MathVista~\cite{lu2023mathvista} & 6,141 & \XSolidBrush & $T, I$ & \XSolidBrush & -- & 83.83 & \XSolidBrush & EN, ZH & SA, OE \\
			GeomVerse~\cite{Geomverse} & 9,303 & \CheckmarkBold & $T, I$ & \CheckmarkBold & 6.35 & 53.09 & \CheckmarkBold & EN & OE \\
			Geo170K~\cite{Gllava} & 177,457 & \CheckmarkBold & $T, I$ & \XSolidBrush & 6.21 & 49.01 & \XSolidBrush & EN & SA \\
			GeoEval~\cite{zhang2024geoeval} & 5,050 & \XSolidBrush & $T, I$ & \CheckmarkBold & 2.55 & 28.34 & \CheckmarkBold & EN & SA, OE \\
			MM-MATH~\cite{sun2024Mmmath} & 5,929 & \XSolidBrush & $T, I$ & \XSolidBrush & 8.50 & 48.50 & \XSolidBrush & EN & OE \\
			AutoGeo-100K~\cite{huang2025autogeo} & 100,000 & \CheckmarkBold & $I$ only & \CheckmarkBold & -- & 22.41 & \CheckmarkBold & EN & -- \\
			MathVerse~\cite{zhang2024mathverse} & 15,672 & \XSolidBrush & $T, I$ & \CheckmarkBold & -- & 40.67 & \XSolidBrush & EN & SA \\
			GeoMM~\cite{R-cot} & 33,595 & \CheckmarkBold & $T, I$ & \CheckmarkBold & 1.51 & 32.47 & \XSolidBrush & EN & OE \\ \hline
			Average & - & - & - & - & 3.92 & 44.98 & - & - & - \\ \hline
			GeoChallenge-90K & 90,279 & \CheckmarkBold & $T, I$ & \CheckmarkBold & 16.72 & 39.15 & \CheckmarkBold & EN, ZH & MA \\ \hline
		\end{tabular}
		\caption{Comparison between GeoChallenge-90K benchmark and existing geometry problem solving benchmarks. AG: Automatic Generation. IF: Input Format, T for text and I for image. CR: Complexity Rating. Avg PL: Average Proof Length. Avg DL: Average Description Length. FA: Formal Annotation. LT: Language Type, EN for English and ZH for Chinese. QT: Question Type, SA for single-answer, MA for multiple-answer and OE for open-ended question}
		\label{table:comparison_with_dataset}
	\end{table*}
	
	\subsubsection{Challenging options generation}
	Given a premise, we enumerate provable conclusions using a symbolic engine, score their difficulty, and select the top-scoring ones as the four options. We use AlphaGeometry~\cite{trinh2024solving} for deduction with a rule set $R$ (theorem matching and algebraic derivation). As summarized in Appendix~\ref{app:forward_chaining}, the engine performs conclusion search via forward chaining, repeatedly applying rules in $R$ to newly derived facts until saturation or a preset depth limit.
	
	

	Following GeoEval~\cite{zhang2024geoeval}, we define difficulty as a weighted sum of five indicators:
	$\text{prior\_difficulty}=\sum_{i=1}^{5} w_i x_i$,
	where $x_1$ is description length, $x_2$ premise length, $x_3$ the number of points, $x_4$ proof-search depth, and $x_5$ proof length. We choose four conclusions with the highest scores, treat one as the correct option, and generate hard distractors via equivalence-preserving rewrites, relation negation (e.g., equality/parallelism), or ratio perturbation, requiring each distractor to be falsifiable under the premise. We avoid naive entity substitution (e.g., $AB\perp CD\rightarrow AB\perp CE$), which often yields degenerate or accidentally true options.

	\subsubsection{Symbol refinement}
	We then refine the symbolic instances and render aligned diagrams. Text refinement includes (i) simplification by retaining only points/clauses used in both the premise and the proofs of candidate options, and (ii) bilingual rendering that maps the formal representation to English and Chinese with rule-based templates while preserving option semantics.
	
	For diagrams, we render figures consistent with the refined premise and explicitly annotate candidate options for visual grounding. To support stable batch rendering beyond the default AlphaGeometry module, we add rule-based option labeling and robustness fixes to prevent incomplete annotations and occasional non-termination.

	\subsubsection{Manual Verification}
	Since symbolic descriptions cannot fully guarantee presentation quality, we perform manual verification as a quality-control filter for the visualizations used in the benchmark. Annotators check (i) \textbf{readability}, legible, unobstructed labels, (ii) \textbf{geometric validity},the drawing satisfies declared relations, and (iii) \textbf{description alignment},required elements present; no contradictory extras. Only figures passing all checks are kept; otherwise they are discarded. This step does not modify the symbolic pipeline and serves purely as visualization quality control.

	\begin{table}[h]
		\centering
		\small
		\begin{tabular}{cccc}
			\hline
			Model & MATH & \begin{tabular}[c]{@{}c@{}}Geome-\\ try3K\end{tabular} & \begin{tabular}[c]{@{}c@{}}GeoCha-\\ llenge\end{tabular} \\ \hline
			Gemini 1.5 Pro & 82.19 & 73.88 & 24.45 \\
			Claude 3.5 Sonnet & 58.05 & 71.65 & 21.81 \\
			GPT-4o & 57.72 & 64.92 & 17.51 \\ \hline
		\end{tabular}
		\caption{Performance comparison of different benchmarks}
		\label{table:comparsion_between_ours_and_others}
	\end{table}
	
	GeoChallenge-90K is designed to evaluate diagram-grounded geometry theorem proving under rigorous, scalable, and controllable settings.
	Built on the fully automatic symbolic pipeline in Section~\ref{section:problem_generation}, all instances are machine-verifiable, enabling scalable construction without manual proof annotation.
	It exhibits six key characteristics: multi-answer MCQ evaluation, automatic generation, comprehensive geometric coverage diversity, dual-modality inputs, bilingual consistency, fine-grained complexity rating.
	Table~\ref{table:comparison_with_dataset} compares GeoChallenge-90K with representative geometry and multimodal math benchmarks.

	\begin{table*}[t]
		\centering
		\small
		\begin{tabular}{lccccccccc}
			\hline
			Model & EMA$\uparrow$ & EME$\uparrow$ & EMM$\uparrow$ & EMH$\uparrow$ & P$\uparrow$ & R$\uparrow$ & F1$\uparrow$ & HA$\uparrow$ & Avg \#Sel \\ \hline
			\multicolumn{10}{l}{\textit{Baselines}} \\
			Random & 6.25 & 6.25 & 6.25 & 6.25 & 50.00 & 50.00 & 50.00 & 46.88 & 2.13 \\ \hline
			\multicolumn{10}{l}{\textit{General-Purpose Models}} \\
			GPT-4o & 17.51 & 29.09 & 13.81 & 9.24 & 52.87 & 68.83 & 59.81 & 58.62 & 2.18 \\
			Claude 3.5 Sonnet & 21.81 & 27.27 & 22.94 & 10.87 & 54.88 & 71.73 & 62.19 & 63.22 & 2.20 \\
			Gemini 1.5 Pro & 24.45 & 38.18 & 22.27 & 9.24 & 62.31 & 76.79 & 68.79 & 69.74 & 2.01 \\
			llava-1.5-7B & 16.96 & 37.82 & 10.69 & 1.09 & 48.31 & 31.36 & 38.03 & 57.87 & 1.03 \\
			Qwen2-VL-7B & 26.65 & 59.64 & 17.37 & 0.0 & 61.36 & 47.44 & 53.51 & 65.69 & 1.19 \\ \hline
			\textit{Reasoning-Oriented Models} &  &  &  &  &  &  &  &  &  \\
			GPT-o3 & 67.84 & 67.27 & 69.04 & 65.76 & 80.84 & 81.20 & 81.02 & 82.52 & 1.64 \\
			GPT-5-nano & 75.89 & 70.59 & 79.31 & 75.00 & 84.82 & 83.78 & 84.30 & 85.49 & 1.61 \\
			Claude 4.5 Sonnet & 41.96 & 50.00 & 41.38 & 30.00 & 72.02 & 77.90 & 74.85 & 76.56 & 1.75 \\
			Gemini 3 Pro & 38.60 & 50.00 & 37.29 & 23.81 & 44.74 & 41.45 & 43.03 & 43.20 & 1.04 \\ \hline
			\textit{Human Performance} &  &  &  &  &  &  &  &  &  \\
			Human & 94.74 & 94.12 & 94.92 & 95.24 & 97.95 & 99.71 & 98.82 & 98.68 & 1.66 \\ \hline
		\end{tabular}
		\caption{Overall performance on GeoChallenge with both text and images provided. EMA/EME/EMM/EMH report Exact Match (EM) on the All/Easy/Medium/Hard splits, respectively.}
		\label{table:main_results}
	\end{table*}

	\subsection{Features of GeoChallenge-90K}
	
	\paragraph{Multi-answer MCQ evaluation.}
	A distinctive feature of GeoChallenge-90K is its multi-answer MCQ format, where an instance may contain more than one correct option. Compared to single-answer MCQs, this setting substantially weakens elimination-style guessing and forces \emph{per-option verification}: models must assess each candidate conclusion under the premise, often requiring different proof paths or relation checks. This format supports option-level metrics and diagnoses over-/under-selection behaviors.

	\paragraph{Long-step Proofs under Concise Statements.} 
	GeoChallenge-90K targets long-step deduction. Proofs average 16.72 steps-over $4\times$ typical geometry benchmarks (3.92)-while statements remain similarly concise (39.15 vs.\ 44.98 words). This is by design that descriptions include only essential relations, increasing information density per token and thus reasoning difficulty. As shown in Table~\ref{table:comparsion_between_ours_and_others}, models suffer a clear accuracy drop on GeoChallenge-90K relative to prior benchmarks, making it a more challenging and more diagnostic testbed.

\begin{figure}[t]
		\centering
		\includegraphics[width=0.95\columnwidth]{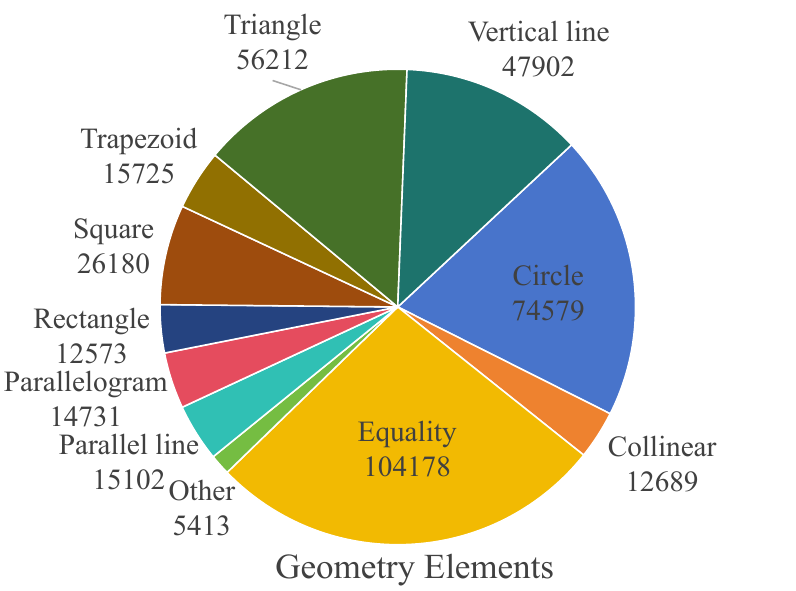}
		\caption{Geometry elements in GeoChallenge-90K}
		\label{Geometry elements}
	\end{figure}

	\paragraph{Comprehensive geometric coverage and structural diversity.}
	Beyond scale, GeoChallenge-90K is constructed to cover a broad range of geometric primitives and composite structures. Figure~\ref{Geometry elements} summarizes the distribution of geometric elements. Importantly, less frequent shapes in many benchmarks, such as trapezoids and parallelograms, are still represented at scale over 10K instances each, which improves coverage of cross-element interactions and mitigates sparsity for rare configurations.
	
	

	\paragraph{Dual-modality and bilingual alignment.} 
	Each instance includes both text and a rendered diagram, and the two modalities are semantically aligned: the textual statement fully specifies the geometric conditions reflected in the diagram. This alignment enables controlled evaluation of text-only versus diagram-grounded reasoning (text+image), and also facilitates vision-only studies when needed. We further provide English and Chinese versions with strict semantic equivalence, reducing confounds from translation artifacts and enabling systematic analysis of language effects.
	
	
	\paragraph{Rich annotations and controlled difficulty.}
	GeoChallenge-90K includes structured formal representations for premises and options, along with two complementary difficulty signals: a prior difficulty estimated from complexity indicators, and posterior difficulty derived from tested models' performance. Problems are stratified into three difficulty levels with a 3:5:2 split, enabling controlled stress testing and fine-grained performance analysis.

	\section{Experiments}

	\subsection{Experimental Setup}

	\paragraph{Prompting / Inference Settings.}
	We adopt a unified prompting protocol across all evaluated models. A single prompt template (included in the supplementary material due to length) is used for every example to elicit a final multiple-choice answer along with brief, option-wise reasoning. We disable external tools and retrieval for all models. Inference is run with greedy decoding (temperature=0.0) and a maximum output length of 16,384 tokens; consequently, each instance is evaluated with one generation only, without additional sampling or self-consistency aggregation.
	
	\paragraph{Models Evaluation.}
	
	To establish a human baseline, we recruited two graduate-level testers with relevant mathematical background. Each problem was translated into their native language, and testers answered under a strict 3-minute limit. Notably, since many items admit correct responses via direct diagram interpretation without a fully formal derivation, the reported human score may reflects a mixture of visual judgment and mathematical reasoning under the same constraint.
	
	We evaluate prominent open-source and closed-source large models selected for strong mathematical reasoning, covering (i) general-purpose LMM/LLMs and (ii) reasoning-oriented models that allocate extra computation for deliberate reasoning. In the multimodal setting, we test GPT-4o~\cite{hurst2024gpt}, Claude 3.5 Sonnet~\cite{anthropic2024claude3.5sonnet}, Gemini 1.5 Pro~\cite{team2024gemini}, LLaVA-1.5-7B~\cite{liu2024llava}, and Qwen2-VL-7B-Instruct~\cite{yang2024qwen2technicalreport}, as well as reasoning-oriented models including GPT-o3-2025-04-16~\cite{openai2025gpto3ando4mini}, GPT-o3-mini-2025-01-31~\cite{openai2025o3mini}, Claude 4.5 sonnet-20250929-thinking~\cite{Anthropic2025claudesonnet45}, and Gemini-3-pro-preview-11-2025~\cite{google2025gemini3pro}. For text-only evaluations (LLM setting), we assess the same model families without visual inputs, and additionally include Deepseek-r1-0528~\cite{deepseekai2025deepseekr1}, Qwen3-235b-a22b-thinking-2507~\cite{yang2025qwen3technicalreport}, and WizardMath-7B~\cite{luo2023wizardmath}.

	For replicability and efficiency, we run most evaluations on GeoChallenge-small, a standardized subset of 908 problems sampled to match the difficulty profile of the full benchmark, enabling practical assessment of closed-source models.
	

	\begin{table*}[t]
		\centering
		\small
		\begin{tabular}{lccccccccc}
			\hline
			Model & EMA$\uparrow$ & EME$\uparrow$ & EMM$\uparrow$ & EMH$\uparrow$ & P$\uparrow$ & R$\uparrow$ & F1$\uparrow$ & HA$\uparrow$ & Avg \#Sel \\ \hline
			\multicolumn{10}{l}{\textit{Baselines}} \\
			Random & 6.25 & 6.25 & 6.25 & 6.25 & 50.00 & 50.00 & 50.00 & 46.88 & 2.13 \\ \hline
			\multicolumn{10}{l}{\textit{General-Purpose Models}} \\
			GPT-4o & 13.99 & 32.00 & 8.69 & 0.0 & 43.75 & 55.89 & 49.08 & 48.62 & 2.14 \\
			Claude 3.5 Sonnet & 17.62 & 32.73 & 15.37 & 0.54 & 51.43 & 63.48 & 56.83 & 60.21 & 2.02 \\
			Gemini 1.5 Pro & 20.59 & 44.00 & 14.70 & 0.0 & 60.32 & 74.14 & 66.52 & 66.77 & 1.97 \\
			WizardMath-7B & 8.15 & 9.45 & 7.80 & 7.07 & 38.48 & 59.32 & 46.68 & 49.78 & 2.40 \\ \hline
			\multicolumn{10}{l}{\textit{Reasoning-Oriented Models}} \\
			GPT-o3-mini & 67.84 & 67.27 & 69.04 & 65.76 & 80.84 & 81.20 & 81.02 & 82.52 & 1.64 \\
			GPT-o3 & 62.67 & 66.18 & 60.80 & 61.96 & 72.95 & 71.41 & 72.17 & 72.60 & 1.59 \\
			Claude 4.5 Sonnet & 36.23 & 51.64 & 34.30 & 17.93 & 66.58 & 67.36 & 66.97 & 71.34 & 1.62 \\
			Gemini 3 Pro & 51.21 & 62.55 & 49.00 & 39.67 & 57.54 & 54.55 & 56.01 & 55.97 & 1.37 \\
			Deepseek r1 & 16.41 & 17.82 & 16.70 & 13.59 & 23.16 & 27.00 & 24.94 & 23.54 & 2.08 \\
			Qwen3-235b-thinking & 75.55 & 81.09 & 75.95 & 66.30 & 82.18 & 82.30 & 82.24 & 82.74 & 1.58 \\ \hline
			\multicolumn{10}{l}{\textit{Human Performance}} \\
			Human & 42.86 & 38.24 & 46.55 & 40.00 & 66.96 & 68.75 & 67.85 & 72.77 & 1.69 \\ \hline
		\end{tabular}
		\caption{Overall performance on GeoChallenge with text-only provided. EMA/EME/EMM/EMH report Exact Match (EM) on the All/Easy/Medium/Hard splits, respectively.}
		\label{table:main_results_text_only}
	\end{table*}

	\paragraph{Evaluation Protocol.}
	We evaluate LLMs on a multi-select benchmark. For each problem $i$ with $K_i$ options, the model predicts a subset $\hat{S}_i \subseteq \{1,\ldots,K_i\}$, compared against the ground-truth subset $S_i$.
	
	We report Exact Match (EM), which is correct iff $\hat{S}_i = S_i$, both overall and by difficulty (Easy/Medium/Hard) determined by our difficulty scores. To measure partial correctness, we compute option-level precision/recall/F1 from the overlap between $\hat{S}_i$ and $S_i$ and report macro-averages across problems. We also report Hamming Loss (HL), the average per-option error rate:
	\[
	\mathrm{HL}=\frac{1}{N}\sum_{i=1}^{N}\frac{1}{K_i}\sum_{j=1}^{K_i}\mathbf{1}\!\left(\hat{y}_{i,j}\neq y_{i,j}\right),
	\]
	where $y_{i,j},\hat{y}_{i,j}\in\{0,1\}$ are the ground-truth and predicted labels; we report Hamming Accuracy (HA) as $1-\mathrm{HL}$. Finally, we report the average number of selected options $\mathbb{E}[|\hat{S}_i|]$ to characterize selection behavior.

	\subsection{Main Results}
	
	We report overall results on GeoChallenge with both text and diagrams provided in Table~\ref{table:main_results}. Since GeoChallenge is formulated as multiple-choice with long-step reasoning, we emphasize Exact Match as the primary metric; in contrast, option-level metrics (P/R/F1 and HA) may still reward partially correct selections even when the predicted answer is not fully correct.
	
	\paragraph{No-guess multiple-choice reveals hidden difficulty.}
	Under the no-guess protocol, the bottleneck shifts from partially identifying correct options to producing a single, answer-consistent prediction. A key observation is the large gap between strict EM and option-level metrics, especially for general-purpose models: e.g., Gemini 1.5 Pro attains high F1/HA but much lower EMA, with similar discrepancies for GPT-4o and Claude 4.5 Sonnet. This indicates that models often recover parts of the correct option answer yet fail to output the exact answer demanded by the no-guess protocol, making heuristics such as eliminating a few options or selecting multiple plausible answers ineffective. Consistently, Avg \#Sel shows that general models select close to two options on average (near the random baseline), trading precision for recall, whereas reasoning-oriented systems are better calibrated and closer to human selection cardinality.
	

	\paragraph{Human–model gap on long-step geometry.}
	GeoChallenge exhibits a substantial human-model gap even with reasoning-oriented systems, and the gap remains on the Hard split. The best model in Table~\ref{table:main_results} (GPT-5-nano) remains obviously below human EMA, and the gap persists on the Hard split; meanwhile, general multimodal baselines lag far behind. This suggests the benchmark separates systems that sustain long-step, answer-consistent reasoning from those that succeed mainly via partial correctness or shallow heuristics. Reasoning-oriented models achieve relatively strong P/R/F1, implying they often locate relevant options, but their remaining deficit under strict EM points to unresolved challenges in multi-step consistency, diagram-grounded verification, and satisfying global geometric constraints.
	
	
	\paragraph{Different degradation with increasing difficulty.}
	As difficulty increases from Easy to Hard, general-purpose models often degrade sharply-sometimes approaching collapse-whereas reasoning-oriented systems decline more moderately and humans remain the most stable. General models typically drop steeply from Easy to Hard, indicating that success on simpler items does not reliably transfer to long-horizon, high-constraint geometric reasoning. In contrast, reasoning-oriented systems better preserve performance as complexity grows. These trends suggest that higher difficulty mainly amplifies failure modes tied to partial-cue reliance, while globally consistent reasoners are more robust under increasing constraints.

	\section{Detailed Findings and Analysis}
	
	\begin{figure*}[h]
		\centering
		\includegraphics[width=0.99\textwidth]{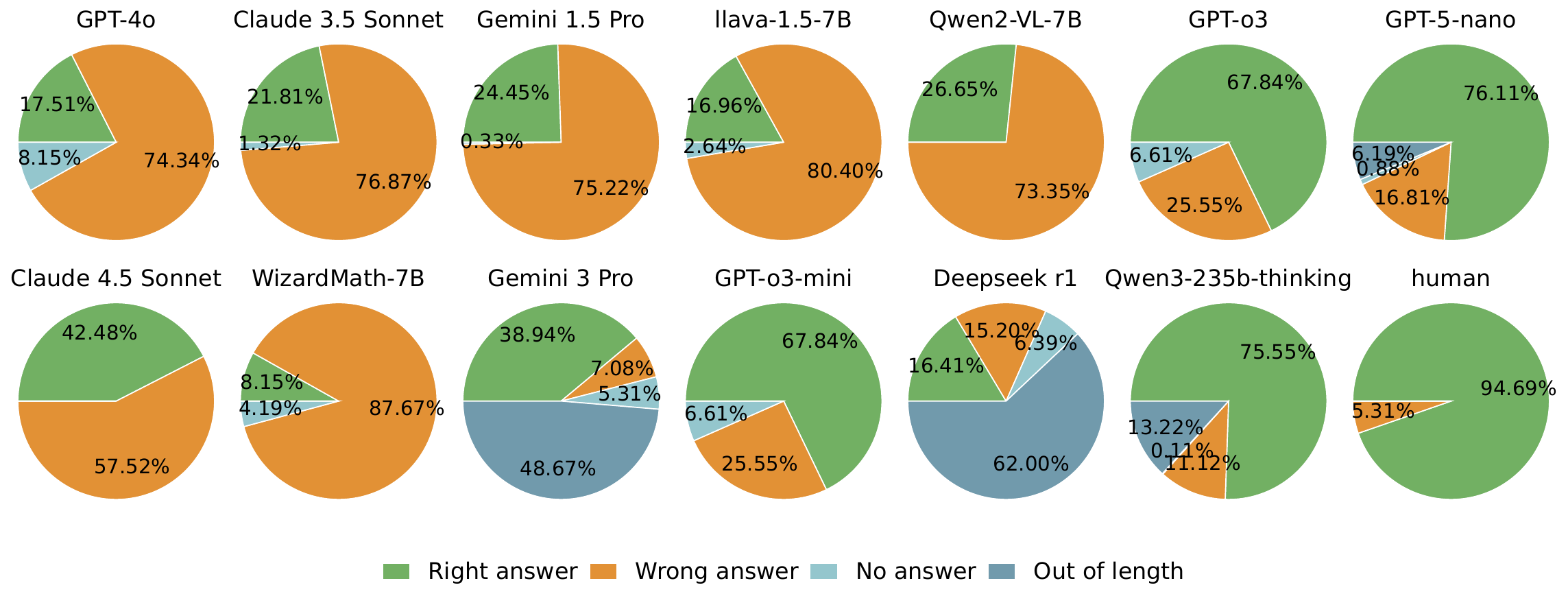}
		\caption{Error type across different models}
		\label{figure:error_type}
	\end{figure*}
	
	We next elaborate on the three findings highlighted in the Abstract and Introduction:
	(1) exact-match failures under the no-guess multi-answer setting,
	(2) weak and inconsistent visual reliance, and
	(3) overextended reasoning without convergence.
	
	\paragraph{Finding 1: Exact-match fragility under no-guess multi-answer MCQ.}
	Under the no-guess protocol, most general-purpose models fail by committing to an incorrect final option answer, suggesting that the primary bottleneck is the answer consistency rather than partial option identification. We categorize each prediction into four mutually exclusive outcomes: \textsc{right\_answer} (exact match), \textsc{wrong\_answer} (committed but incorrect), \textsc{no\_answer} (uncommitted answer), and \textsc{out\_of\_length} (truncation).
	Figure~\ref{figure:error_type} shows that general-purpose models are largely driven by \textsc{wrong\_answer} (roughly three quarters), while \textsc{no\_answer} varies widely across systems, reflecting different uncertainty-handling behaviors. In contrast, reasoning-oriented models exhibit more structured trade-offs (higher \textsc{right\_answer} with mild abstention), motivating us to report error-type composition alongside EM.
	
	\paragraph{Additional evidence: Language shift stresses exact match.}
	
	Strict EMA is more sensitive to English$\rightarrow$Chinese reformulation than option-level metrics, indicating that linguistic changes can break exact answer consistency even when models still identify relevant options. We evaluate general-purpose models under Chinese prompts and compare them with English prompts to assess cross-lingual robustness, and observe a recurring over-selection pattern under Chinese prompts (higher Avg \#Sel): this can maintain or even improve F1/HA, yet lowers EMA because any extra or missed option turns an exact match into zero under the no-guess protocol (see Appendix~\ref{table:geochallenge_Result_in_chinese_input} for detailed results).

	\paragraph{Finding 2: Weak and inconsistent diagram grounding.}
	Humans rely heavily on diagrams, whereas current LLMs under-use or inconsistently integrate visual evidence. To probe diagram reliance, we compare multimodal (text+image) performance against text-only performance, and additionally report a human text-only baseline. Humans show strong dependence on diagrams: removing the diagram causes a 51.88 EMA drop, while option-level quality remains comparatively strong, suggesting that people can partially reconstruct the figure from text and preserve partial correctness even when exact selection becomes harder. In contrast, most LLMs are only weakly diagram-dependent: performance usually declines but not catastrophically, and the benefit of visual input is inconsistent across models; in some cases, text-only can even outperform multimodal. Overall, diagram usage emerges as a key capability dimension: humans treat diagrams as the primary substrate for geometry, whereas current LLMs do not reliably ground deductions in visual evidence.

	\paragraph{Finding 3: Overextended reasoning without convergence.}
	A non-trivial fraction of failures stems from non-convergent long-step reasoning that does not reach a stable final selection within the decoding budget, recorded as \textsc{out\_of\_length}. With \texttt{max\_tokens}=16{,}384, \textsc{out\_of\_length} typically indicates that the model keeps expanding or revising its reasoning without committing to a verifiable answer set, rather than being merely verbose. This failure mode is especially visible in models that attempt prolonged, open-ended deliberation (e.g., DeepSeek-R1 in Figure~\ref{figure:error_type}), and it becomes increasingly harmful under strict EM: even near-correct intermediate judgments are not credited unless the model converges to a final, consistent option set.

	\section{Conclusion}
	In this work, we presented GeoChallenge-90K, a large-scale benchmark for evaluating diagram-grounded geometry theorem proving in a rigorous and scalable setting. Our benchmark is constructed through a symbolic generation-and-verification pipeline, provides aligned text-diagram inputs with bilingual consistency, and exposes controllable complexity for fine-grained analysis. To better reflect real reasoning ability, we further adopt a no-guess, multi-answer multiple-choice protocol that enables strict exact-match evaluation while still allowing complementary option-level diagnostics.
	
	Across comprehensive experiments and analyses, GeoChallenge-90K consistently separates shallow pattern matching from long-step, globally consistent reasoning: current general-purpose models struggle under strict evaluation, reasoning-oriented models improve substantially yet remain behind human solvers, and the gap widens as complexity increases. Our diagnostic studies suggest that the remaining challenges are not merely harder problems, but failures in answer consistency, convergence, and reliable integration of diagram evidence. We hope GeoChallenge-90K will serve as a practical testbed for developing and measuring future systems that reason over long proofs, verify global constraints, and ground deductions in diagrams more reliably.

	\section{Limitations}
	
	Our work has three main limitations. First, while GeoChallenge-90K supports rigorous outcome-level evaluation (e.g., strict exact match, option-level metrics, and error-type composition), we do not conduct fine-grained, step-by-step process analyses to localize errors to specific intermediate decisions or proof-state transitions.Second, although complexity control and text-only ablations are useful diagnostics, they do not causally identify why models fail. They suggest weaknesses in visual grounding and long-step consistency, but cannot pinpoint which visual cues (e.g., annotations or intersections) or reasoning operations (e.g., parsing or constraint propagation) are the primary error sources; targeted perturbation studies would be needed.Third, some failure modes-notably \textsc{no\_answer} and \textsc{out\_of\_length}-are sensitive to prompting and decoding choices. Variations in answer-format constraints, refusal behavior, or search strategies can shift abstention and non-convergence rates, affecting error-type composition even when underlying competence is similar.

	\bibliography{custom}
	
	\appendix
	\section{Examples of Clauses, Constructions, and Premises}
	\label{sec:reference_examples}
	\paragraph{Clauses}
	
	$triangle$: This clause created three points to construct a triangle.
	
	$midpoint\ a\ b$: This clause uniquely determines a point based on the positions of points A and B.
	
	$angle\_bisector\ a\ b\ c$: This clause describes the set of points that lie in the angle bisector of $\angle ABC$.
	
	\paragraph{Constructions}
	
	$a\ b\ c\ =\ triangle$: This construction created three points to construct a triangle.
	
	$x\ =\ angle\_bisector\ a\ b\ c,\ on\_line\ a\ c$: This construction uniquely determines the intersection point between the angle bisector of $\angle ABC$ and the line AC.
	
	$x\ =\ midpoint\ a\ b,\ midpoint\ a\ c$(invalid): This is an invalid construction, as the midpoints of AB and AC generally do not intersect at a single, well-defined point.
	
\begin{figure}[h]
		\centering
		\includegraphics[width=0.60\columnwidth]{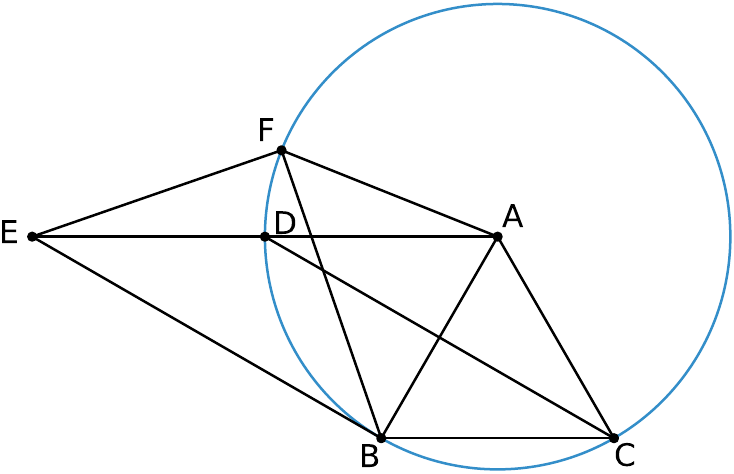}
		\caption{the diagram corresponding to the premise above}
		\label{figure:800023_pic}
	\end{figure}

	\paragraph{Premise}
	\begin{align*}
		a\,b\,c &= \mathrm{ieq\_triangle}(a,b,c);\\
		d &= \mathrm{reflect}(d,c,a,b);\\
		e &= \mathrm{on\_line}(e,d,a),\ \mathrm{eqdistance}(e,d,a,b);\\
		f &= \mathrm{on\_dia}(f,b,e),\ \mathrm{on\_circle}(f,a,c).
	\end{align*}
	This premise captures six point elements and their geometric relationships, which enables the construction of a diagram that matches the description (Figure \ref{figure:800023_pic})
	
	\section{Forward Chaining Algorithm}
	~\label{app:forward_chaining}
	This section details our forward-chaining conclusion search strategy: starting from the premise, we iteratively match theorems and derive algebraic facts under a maximum depth $N$, accumulating all newly derived conclusions until no further updates occur.
	
	\begin{algorithm}[tb]
		\caption{Conclusion search via forward chaining}
		\label{alg:ag_ddar}
		\textbf{Input}: Premise\\
		\textbf{Parameter}: max\_level $N$\\
		\textbf{Output}: All derived conclusions\par
		\begin{algorithmic}[1]
			\STATE $C \gets$ Read(Premise)
			\STATE $level \gets 0$
			\WHILE{$level < N$}
			\STATE $T \gets$ match\_theorems(R)
			\STATE $T \gets T \cup$ derive\_algebra(R)
			\IF{Not\_empty($T$)}
			\STATE $C \gets C \cup T$
			\ELSE
			\STATE \textbf{break}
			\ENDIF
			\ENDWHILE
			\STATE \textbf{return} $C$
		\end{algorithmic}
	\end{algorithm}
	
	\section{GeoChallenge Details}
	\label{appendix:geochallengedetails}
	
	\begin{table*}[t]
		\small
		\centering
		\begin{tabular}{lccccccccc}
			\hline
			Model & EMA$\uparrow$ & EME$\uparrow$ & EMM$\uparrow$ & EMH$\uparrow$ & P$\uparrow$ & R$\uparrow$ & F1$\uparrow$ & HA$\uparrow$ & Avg \#Sel \\ \hline
			\multicolumn{10}{l}{\textit{Both text and images provided}} \\
			GPT-4o & 11.89 & 19.64 & 9.13 & 7.07 & 40.27 & 54.66 & 46.38 & 45.84 & 2.25 \\
			Claude 3.5 Sonnet & 17.62 & 30.18 & 13.59 & 8.70 & 52.37 & 64.86 & 57.95 & 60.85 & 2.05 \\
			Gemini 1.5 Pro & 24.23 & 38.55 & 21.16 & 10.33 & 62.85 & 82.98 & 71.52 & 71.31 & 2.12 \\
			llava-1.5-7B & 12.11 & 25.09 & 8.24 & 2.17 & 37.00 & 23.76 & 28.94 & 50.61 & 1.05 \\
			Qwen2-VL-7B & 24.01 & 58.91 & 12.47 & 0.0 & 57.16 & 48.97 & 52.75 & 63.13 & 1.41 \\ \hline
			\multicolumn{10}{l}{\textit{Text-only provided}} \\
			GPT-4o & 9.58 & 19.27 & 7.57 & 0.0 & 43.45 & 68.81 & 53.26 & 46.72 & 2.71 \\
			Claude 3.5 Sonnet & 12.78 & 23.64 & 11.36 & 0.0 & 48.39 & 69.37 & 57.02 & 57.74 & 2.34 \\
			Gemini 1.5 Pro & 23.35 & 54.91 & 13.59 & 0.0 & 63.42 & 74.08 & 68.34 & 68.72 & 1.86 \\
			WizardMath-7B & 7.71 & 10.18 & 7.13 & 5.43 & 38.34 & 48.55 & 42.85 & 47.58 & 2.07 \\ \hline
		\end{tabular}
		\caption{General-purpose models' performance on GeoChallenge, where the same problems are presented with Chinese prompts rather than English.}
		\label{table:geochallenge_Result_in_chinese_input}
	\end{table*}
	
	The GeoChallenge-90K dataset comprises 90,279 automatically generated geometric proof problems. As detailed in Table \ref{table:details} of the Appendix \ref{appendix:geochallengedetails}, all problems presented in our dataset are offered with images and solutions of the problem, with both English and Chinese version of the problem provided. the dataset exhibits a stratified difficulty distribution with 27,083 easy-level (30.0\%), 45,140 medium-level (50.0\%), and 18,056 hard-level (20.0\%) problems. This partitioning follows a 3:5:2 ratio based on a priori complexity scores derived from the weighted function shown in main text.
	
	\begin{table}[h]
		\centering
		\begin{tabular}{lc}
			\hline
			\multicolumn{2}{c}{\textbf{Statistics}} \\ \hline
			Total Problems & 90,279 \\ \hline
			\begin{tabular}[c]{@{}l@{}}Coverage\\ (Img / Sol / EN / ZH)\end{tabular} & \begin{tabular}[c]{@{}c@{}}100\% / 100\% /\\ 100\% / 100\%\end{tabular} \\ \hline
			Difficulty & 90,279 \\
			\textit{* Easy} & 27,083 (30.0\%) \\
			\textit{* Medium} & 45,140 (50.0\%) \\
			\textit{* Hard} & 18,056 (20.0\%) \\ \hline
			Geometry Relations & 258,963 \\
			\textit{* Equality} & 104,178 \\
			\textit{* Parallel} & 15,012 \\
			\textit{* Vertical} & 47,092 \\
			\textit{* Collinear} & 12,689 \\
			\textit{* Circle-related} & 74,579 \\
			\textit{* Others} & 5,413 \\ \hline
		\end{tabular}
		\caption{Detailed statistics of the GeoChallenge dataset. EN and ZH denote English and Chinese statements, respectively.}
		\label{table:details}
	\end{table}

	
	\section{Predefined Clause Templates}
	
	This appendix summarizes the predefined clause templates used in our automatic geometry-problem generation pipeline; the complete list is provided in Table \ref{table:defs_template}.
	
	\begin{table*}[t]
		\centering
		\begin{tabular}{ll}
			\hline
			\multicolumn{1}{c}{name} & \multicolumn{1}{c}{meaning} \\ \hline
			X = angle\_bisector(A,B,C) & Construct point X on the angle bisector of $\angle$ ABC \\
			X = angle\_mirror(A,B,C) & Construct point X such that BC is the bisector of $\angle$ ABX \\
			X,Y,Z,I = centroid(A,B,C) & Construct point I as the centroid of $\triangle$ABC with midpoints X, Y, Z \\
			X = circle(A,B,C) & Construct point X as the circumcenter of $\triangle$ABC \\
			A,B,C,D = eq\_quadrangle() & Construct quadrilateral ABCD with AD = BC \\
			A,B,C,D = eq\_trapezoid() & Construct trapezoid ABCD with AD = BC \\
			X = eq\_triangle(B,C) & Construct point X such that $\triangle$XBC is an equilateral triangle \\
			X = eqangle2(A,B,C) & Construct point X such that $\angle$ BAX = $\angle$ XCB \\
			X = eqangle3(A,B,D,E,F) & Construct point X such that $\angle$ AXB = $\angle$ EDF \\
			A,B,C,D = eqdia\_quadrangle() & Construct quadrilateral ABCD with AC = BD \\
			X = eqdistance(A,B,C) & Construct point X such that XA = BC \\
			X = excenter(A,B,C) & Construct point X as the excenter of $\triangle$ABC \\
			X = foot(A,B,C) & Construct point X as the foot of A on BC \\
			A,B,C = ieq\_triangle() & Construct equilateral triangle $\triangle$ABC\\
			X = incenter(A,B,C) & Construct point X as the incenter of $\triangle$ABC \\
			X = intersect(f,g) & Construct point X as function f() $\cap$ g() \\
			X = intersection\_lc(A,O,B) & Construct point X as circle O $\cap$ line AB \\
			X = intersection\_ll(A,B,C,D) & Construct point X as line AB $\cap$ line CD \\
			X = intersection\_lp(A,B,C,M,N) & Construct point X as line AB $\cap$ line through C parallel to line MN \\
			X = intersection\_lt(A,B,C,D,E) & Construct point X as line AB $\cap$ line through C perpendicular to line DE \\
			X = intersection\_pp(A,B,C,D,E,F) & Construct point X such that line XA $\parallel$ line BC and line XD $\parallel$ line EF \\
			X = intersection\_tt(A,B,C,D,E,F) & Construct point X such that line XA $\perp$ line BC and line XD $\perp$ line EF \\
			A,B,C = iso\_triangle() & Construct A, B, C such that AB = AC \\
			A,B,C,D = isquare() & Construct square ABCD \\
			X = lc\_tangent(A,O) & Construct point X such that OA is perpendicular to AX \\
			X = midpoint(A,B) & Construct point X as the midpoint of AB \\
			X = mirror(A,B) & Construct point X such that B is the midpoint of AX \\
			X,Y,Z,I = ninepoints(A,B,C) & Construct midpoints X, Y, Z and point I as the circumcenter of $\triangle$XYZ \\
			X = nsquare(A,B) & Construct point X such that XB is the right isosceles triangle \\
			X = on\_aline(A,B,C,D,E) & Construct point X such that $\angle$ XAB = $\angle$ CDE \\
			X = on\_bline(X,A,B) & Construct point X on the perpendicular bisector of AB \\
			X = on\_ circle(O,A) & Construct point X such that OA = OX \\
			X = on\_circum(A,B,C) & Construct point X on the circumcircle of A, B, C \\
			X = on\_dia(A,B) & Construct point X such that AX is perpendicular to BX \\
			X = on\_line(A,B) & Construct point X on line AB \\
			X = on\_pline(A,B,C) & Construct point X such that XA is parallel to BC \\
			X = on\_tline(A,B,C) & Construct point X such that XA is perpendicular to BC \\
			X = orthocenter(A,B,C) & Construct point X as the orthocenter of ABC \\
			X = parallelogram(A,B,C) & Construct point X such that ABCX is a parallelogram \\
			A,B,C,D = r\_trapezoid() & Construct right trapezoid ABCD \\
			A,B,C = r\_triangle() & Construct right triangle ABC \\
			A,B,C,D = rectangle() & Construct rectangle ABCD \\
			X = reflect(A,B,C) & Construct point X as the reflection of A about BC \\
			A,B,C = risos() & Construct point X as the isosceles triangle ABC \\
			X = s\_angle(A,B,$\alpha$) & Construct point X such that $\angle$ ABX = $\alpha$ \\
			A,B = segment() & Construct segment AB \\
			X = shift(B,C,D) & Construct point X such that XB=CD and XC=BD \\
			X Y = square(A,B) & Construct point X, Y such that XYAB is a square \\
			X,Y = tangent(A,O,B) & Construct point X,Y as the tangent touch points from A to circle (O,B) \\
			A,B,C,D = trapezoid() & Construct trapezoid ABCD \\
			A,B,C = triangle() & Construct triangle ABC \\ \hline
		\end{tabular}
		\label{table:defs_template}
		\caption{Predefined Clause Templates for Generating Geometric Problems (Adapted and modified from Work AlphaGeometry) }
		\label{table:defs_template}
	\end{table*}
	
	\section{Predefined Rules}
	
	This appendix lists the predefined inference rules employed by the symbolic reasoning engine for forward deduction; the full rule set is presented in Table \ref{table:rules_template}.
	
	\begin{table*}[t]
		\centering
		\begin{tabular}{ll}
			\hline
			\multicolumn{1}{c}{premise(s)} & \multicolumn{1}{c}{conclusion} \\ \hline
			perp A B C D, perp C D E F, ncoll A B E & para A B E F \\
			cong O A O B, cong O B O C, cong O C O D & cyclic A B C D \\
			eqangle A B P Q C D P Q & para A B C D \\
			cyclic A B P Q & eqangle P A P B Q A Q B \\
			eqangle6 P A P B Q A Q B, ncoll P Q A B & cyclic A B P Q \\
			cyclic A B C P Q R, eqangle C A C B R P R Q & cong A B P Q \\
			midp E A B, midp F A C & para E F B C \\
			midp E A B, midp F A C, midp G B C & cong E F G B \\
			midp E A B, midp F A C, midp G B C & cong E F G C \\
			para A B C D, coll O A C, coll O B D & eqratio3 A B C D O O \\
			perp A B C D, perp E F G H, npara A B E F & eqangle A B E F C D G H \\
			eqangle A B C D M N P Q, eqangle C D E F P Q R U & eqangle A B E F M N R U \\
			eqratio A B C D M N P Q, eqratio C D E F P Q R U & eqratio A B E F M N R U \\
			eqratio6 D B D C A B A C, Coll D B C, ncoll A B C & eqangle6 A B A D A D A C \\
			eqangle6 A B A D A D A C, Coll D B C, ncoll A B C & eqratio6 D B D C A B A C \\
			cong O A O B, ncoll O A B & eqangle O A A B A B O B \\
			eqangle6 A O A B B A B O, ncoll O A B & cong O A O B \\
			circle O A B C, perp O A A X & eqangle A X A B C A C B \\
			circle O A B C, eqangle A X A B C A C B & perp O A A X \\
			circle O A B C, midp M B C & eqangle A B A C O B O M \\
			\begin{tabular}[c]{@{}l@{}}circle O A B C, coll M B C,\\ eqangle A B A C O B O M\end{tabular} & midp M B C \\
			perp A B B C, midp M A C & cong A M B M \\
			circle O A B C, coll O A C & perp A B B C \\
			cyclic A B C D, para A B C D & eqangle A D C D C D C B \\
			midp M A B, perp O M A B & cong O A O B \\
			cong A P B P, cong A Q B Q & perp A B P Q \\
			cong A P B P, cong A Q B Q, cyclic A B P Q & perp P A A Q \\
			midp M A B, midp M C D & para A C B D \\
			midp M A B, para A C B D, para A D B C & midp M C D \\
			\begin{tabular}[c]{@{}l@{}}eqratio O A A C O B B D, coll O A C,\\ coll O B D, ncoll A B C, sameside A O C B O D\end{tabular} & para A B C D \\
			para A B A C & coll A B C \\
			midp M A B, midp N C D & eqratio M A A B N C C D \\
			eqangle A B P Q C D U V, perp P Q U V & perp A B C D \\
			eqratio A B P Q C D U V, cong P Q U V & cong A B C D \\
			\hline
		\end{tabular}
		\caption{Rules Used by the Symbolic Reasoning Engine for Deriving New Conclusions}
		\label{table:rules_template}
	\end{table*}
	
	\section{Cross-lingual robustness under Chinese prompts}
	
	We evaluate general-purpose models under Chinese prompts and compare them with their English-prompt counterparts to assess cross-lingual robustness, reporting strict exact-match accuracy (EMA) alongside option-level metrics (F1, HA) and the average number of selected options (Avg \#Sel). Across models, EMA is consistently more sensitive to language shifts than option-level metrics, indicating that changes in linguistic formulation can disrupt exact answer consistency even when partial option identification remains similar.
	
	As shown in table \ref{table:geochallenge_Result_in_chinese_input}, in the multimodal setting (text + images), GPT-4o degrades substantially from English to Chinese (EMA -5.62, F1 -13.43, HA -12.78), and Claude 3.5 Sonnet also drops consistently (EMA -4.19, F1 -4.24, HA -2.37), whereas Gemini 1.5 Pro remains largely stable (EMA -0.22) with slightly improved F1/HA, suggesting stronger cross-lingual robustness. We also observe a recurring over-selection pattern under Chinese prompts: Qwen2-VL-7B shows a smaller EMA decrease (26.65 to 24.01) but selects more options (+0.22 Avg \#Sel), and under text-only evaluation GPT-4o exhibits the same mechanism more clearly (EMA -4.41 vs. F1 +4.18 with Avg \#Sel +0.57). Taken together, language-induced over-selection can preserve or even improve F1/HA by increasing coverage of correct options, while reducing EMA because any extra or missing option breaks the exact match.
	
	\section{LLM Prompt Templates}
	
	All model evaluations were conducted using carefully designed and standardized system prompts to ensure full reproducibility across different experimental settings. The prompt templates remained consistent throughout all evaluation scenarios, with minor adaptations made only for modality-specific requirements.

	For text-only model evaluations, we maintained strict prompt consistency by using identical textual templates across all comparable experiments. Notably, despite the visual nature of some tasks, we intentionally omitted any special instructions regarding missing images in text-only settings. This design choice was based on two key considerations: (1) our problem descriptions maintain high text-image consistency, allowing models to theoretically reconstruct the visual information from textual descriptions alone (though this capability may be challenging for current models), and (2) we observed that models typically wouldn't refuse to answer due to absent images when provided with sufficiently detailed textual descriptions.

\begin{figure*}[t]
		\centering
		\includegraphics[width=0.99\textwidth]{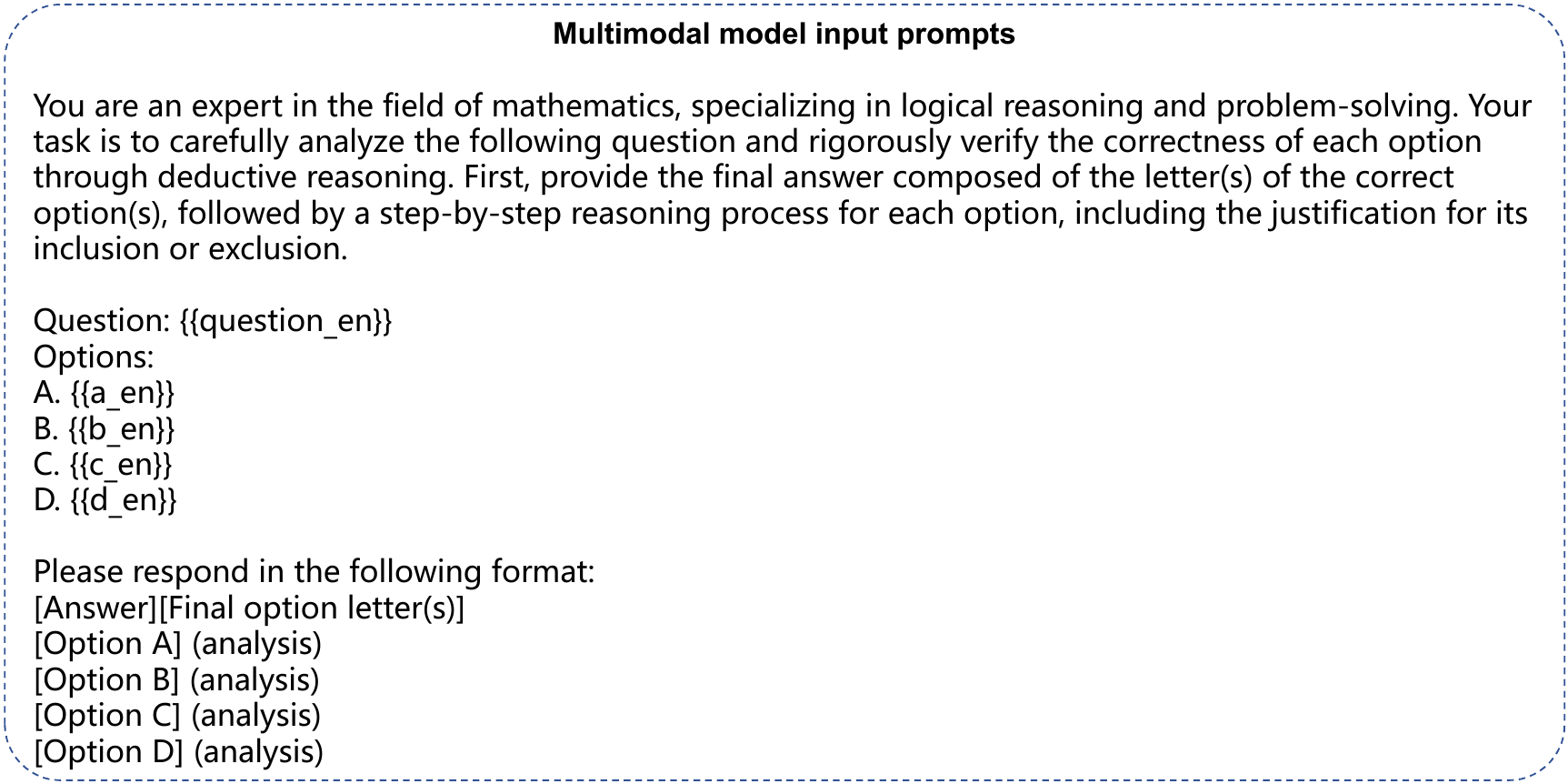}
		\caption{English version of prompt to LLMs}
		\label{figure:en_prompt}
	\end{figure*}

\begin{figure*}[t]
		\centering
		\includegraphics[width=0.99\textwidth]{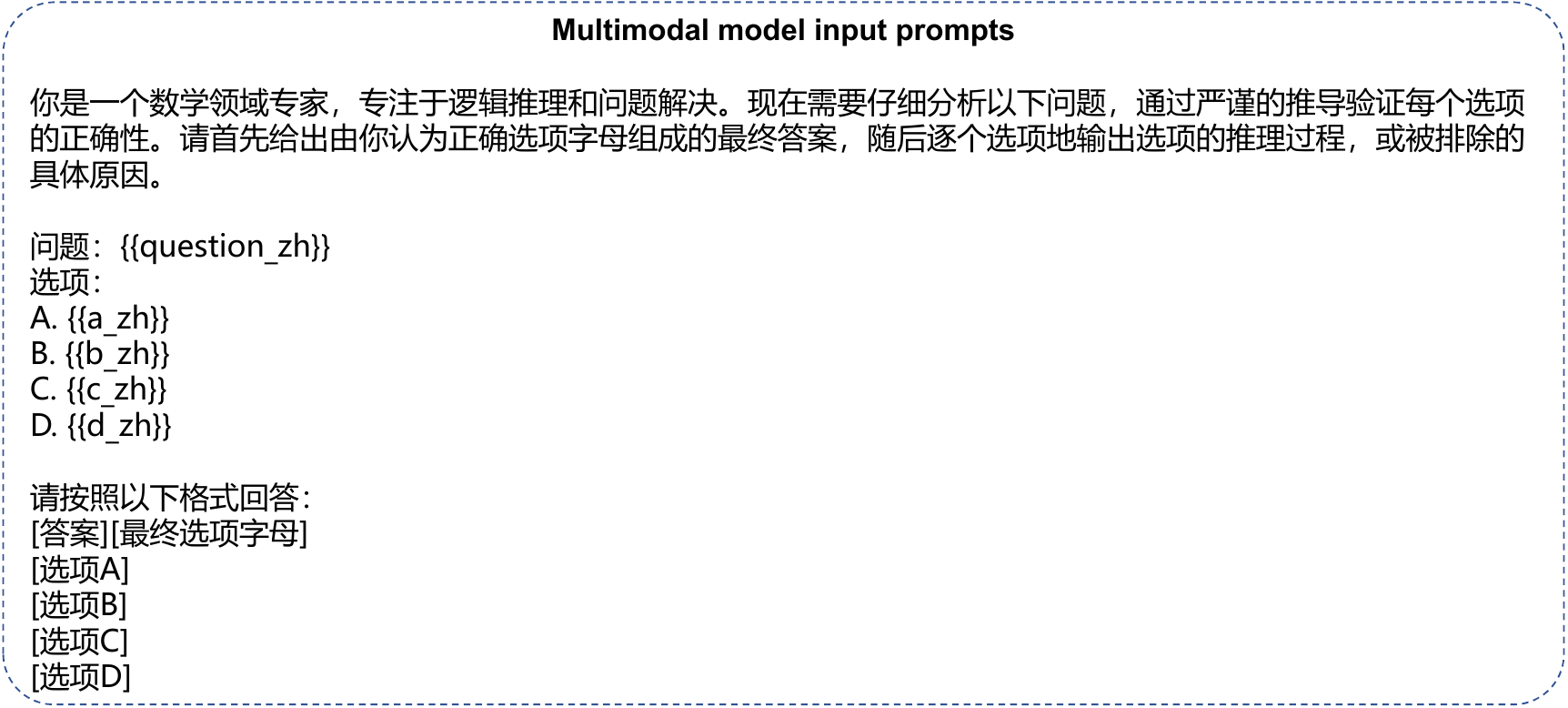}
		\caption{Chinese version of prompt to LLMs}
		\label{figure:zh_prompt}
	\end{figure*}

	For text+image evaluations, we employed the structured prompt templates illustrated in Figure \ref{figure:en_prompt} and \ref{figure:zh_prompt} for English and Chinese versions respectively. These multimodal prompts systematically incorporated both the textual instructions and visual content, with clear markers distinguishing between image inputs and textual components. The visual examples accompanying each prompt were carefully selected to be representative of the task requirements while avoiding potential biases in image content or composition.

	\section{More Problems in GeoChallenge Dataset}
	
\begin{figure*}[t]
		\centering
		\includegraphics[width=0.99\textwidth]{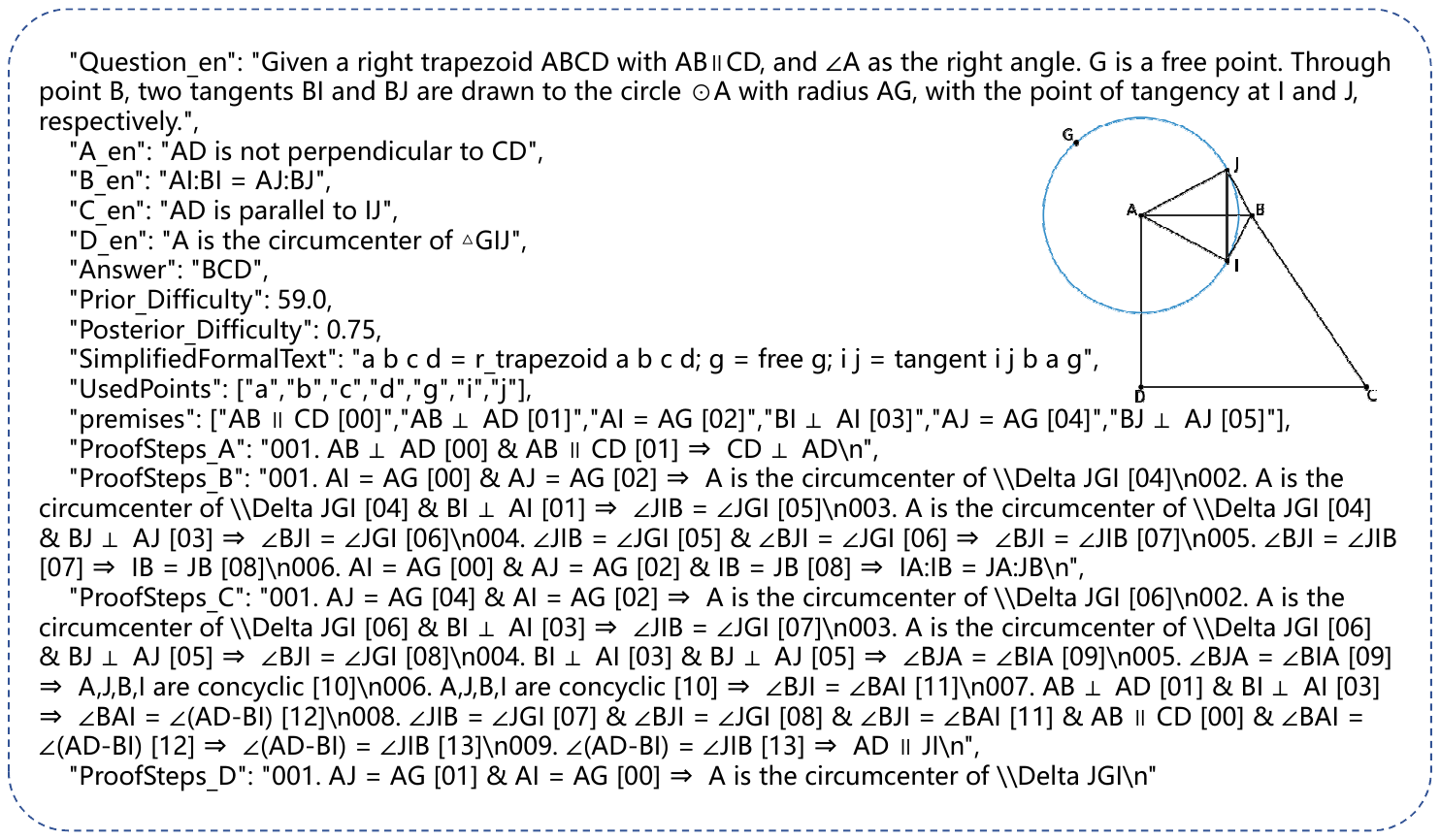}
		\caption{Example of easy level problem in GeoChallenge}
		\label{figure:easy}
	\end{figure*}

\begin{figure*}[t]
		\centering
		\includegraphics[width=0.99\textwidth]{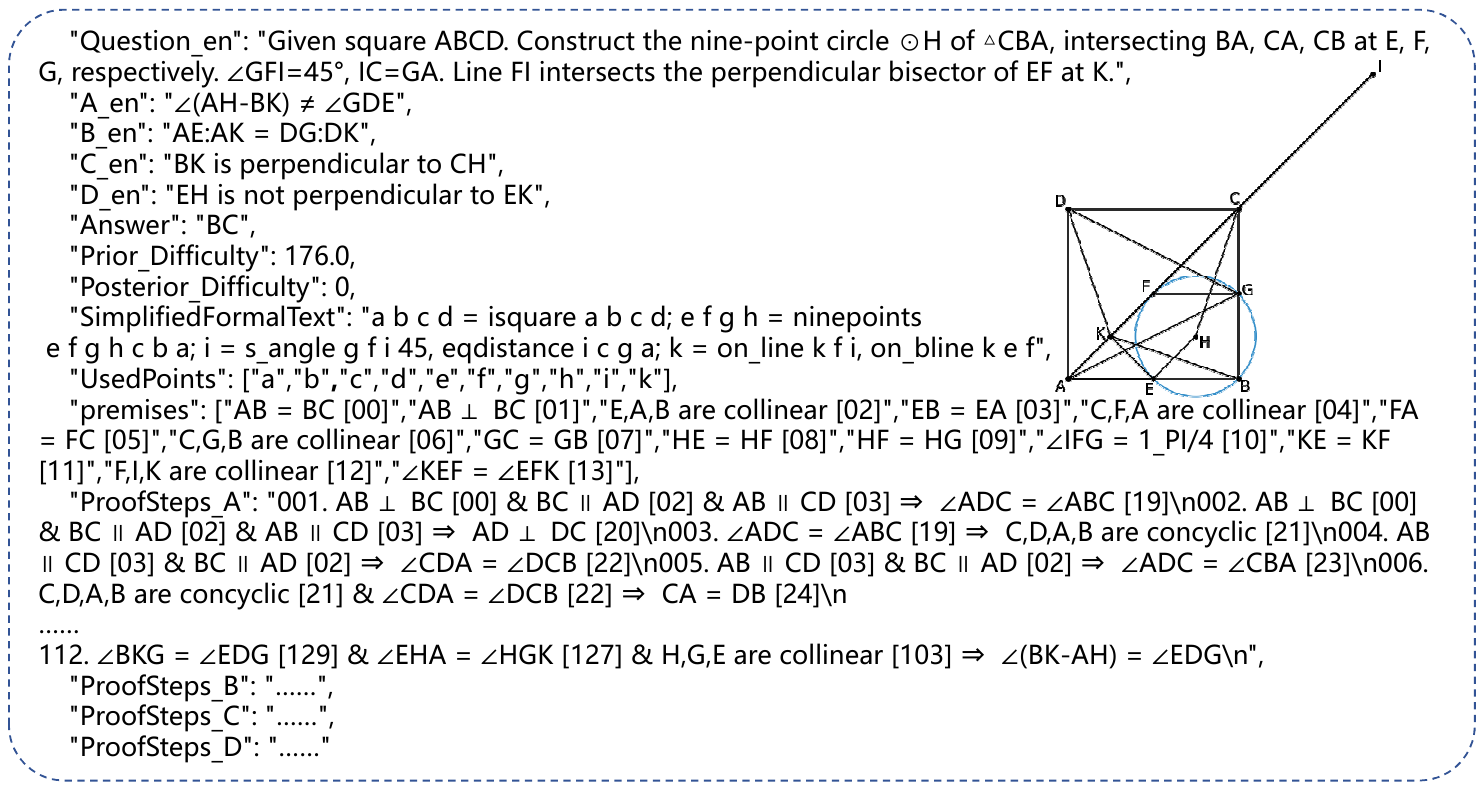}
		\caption{Example of medium level problem in GeoChallenge}
		\label{figure:medium}
	\end{figure*}

\begin{figure*}[t]
		\centering
		\includegraphics[width=0.99\textwidth]{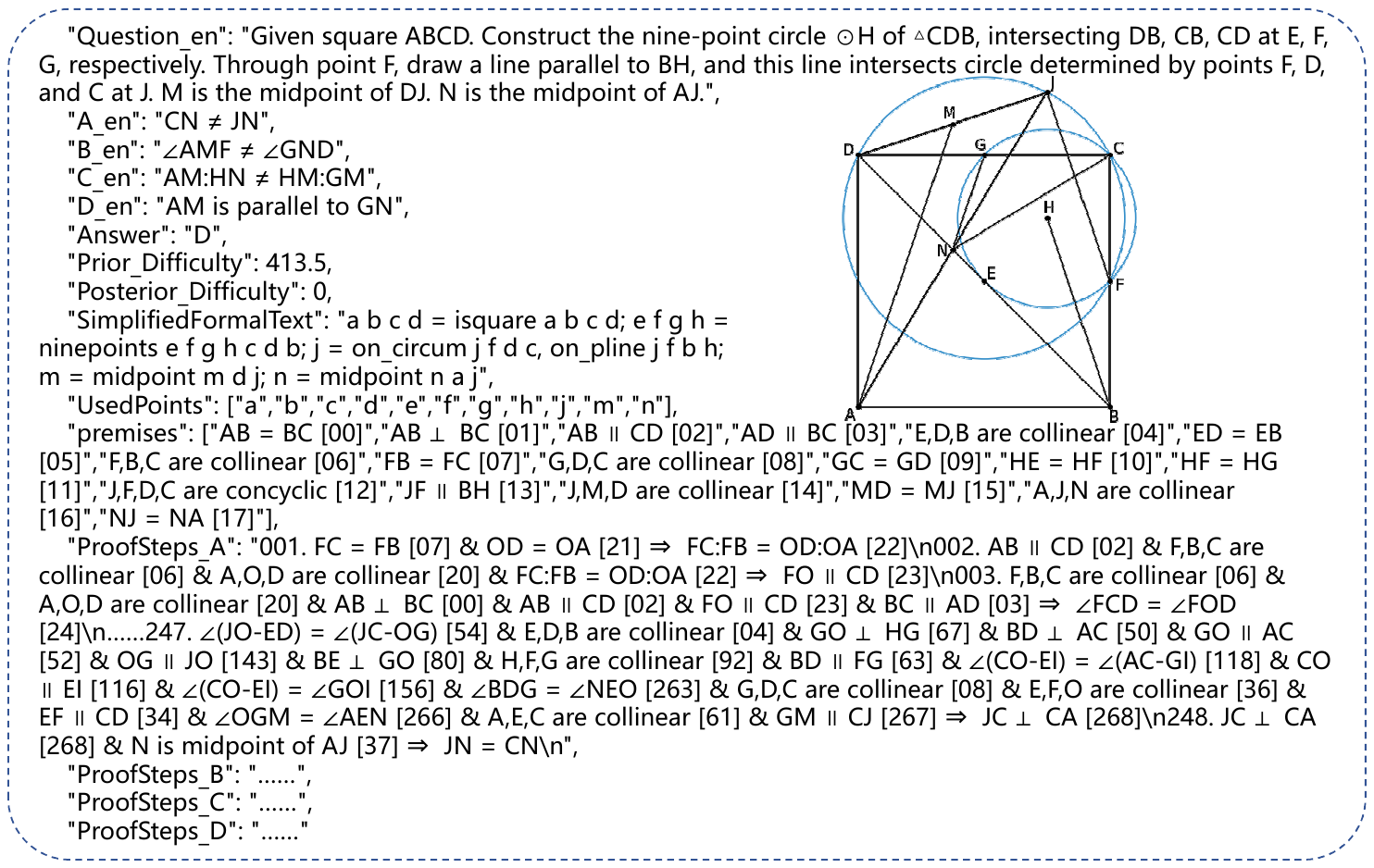}
		\caption{Example of hard level problem in GeoChallenge}
		\label{figure:hard}
	\end{figure*}

	Figures \ref{figure:easy}, \ref{figure:medium}, and \ref{figure:hard} illustrate representative examples of easy, medium, and hard problems, respectively. A clear progression in complexity is observable across these difficulty levels: the textual descriptions grow significantly longer, incorporating more intricate logical constraints and nuanced phrasing, while the accompanying images exhibit a marked increase in structural sophistication, evidenced by the rising number of geometric points, connecting lines, and layered annotations. Notably, the easy-level problems typically involve straightforward deductions with minimal intermediate reasoning steps, whereas medium and hard problems demand deeper theorem applications, often requiring multi-hop inference chains and careful consideration of implicit spatial relationships. This escalation in cognitive demand aligns closely with human intuition—the harder problems not only present more visual clutter but also necessitate greater mental effort in parsing, planning, and executing solutions. The deliberate stratification of difficulty ensures that the benchmark captures a wide spectrum of reasoning capabilities, from basic pattern recognition to advanced geometric theorem synthesis, mirroring the gradual skill development observed in human problem-solving. Furthermore, the consistency between objective complexity metrics (e.g., token count, graph density) and subjective human assessment underscores the validity of our difficulty calibration methodology.  
	
\end{document}